\documentclass[10pt,journal]{IEEEtran}
\usepackage{amsmath,amsfonts,epsfig,dcolumn,bm,subfigure,version}
\usepackage[textwidth=2cm, textsize=small]{todonotes}

\def\MFCCDD{MFCC+$\Delta$+$\Delta\Delta$}
\def\PLPDD{PLP+$\Delta$+$\Delta\Delta$}
\def\DD{$\Delta$+$\Delta\Delta$}

\def\Z{{\relax\ifmmode Z\!\!\!Z\else$Z\!\!\!Z$\fi}}
\def\R{{\relax\ifmmode I\!\!R\else$I\!\!R$\fi}}

\newcommand{\figwidth}{3.4in}
\begin{document}
\title{Speech Recognition Front End Without Information Loss }
\author{Matthew~Ager, Zoran~Cvetkovi\'c~\IEEEmembership{Senior Member,~IEEE,} and Peter~Sollich%
\thanks{This work was done while M.\ Ager was with the Department of Mathematics, King's College London. P.\ Sollich is with the Department of Mathematics and Z.\ Cvetkovi\'c is with the Department of
Informatics, King's College London, Strand, London WC2R 2LS, UK.}%
\thanks{This project was supported by EPSRC Grant EP/D053005/1.}
\thanks{This work was presented in part at ISIT 2011~\cite{isit11}.}}%

\maketitle

\begin{abstract}
Speech representation and modelling in high-dimensional spaces of acoustic waveforms, or a linear transformation thereof,  is investigated 
with the aim of improving the robustness of automatic speech recognition to additive noise. The motivation behind this approach is twofold: (i) the information in acoustic waveforms  that is 
usually removed in the process of extracting low-dimensional features might aid robust recognition by virtue of structured redundancy analogous to channel coding, (ii)  linear feature domains allow for exact noise adaptation,
as opposed to representations that involve non-linear processing which makes noise adaptation challenging. Thus,
we develop a generative framework for 
phoneme modelling in high-dimensional  linear feature domains, and use it  in phoneme classification and recognition tasks.
Results show that classification and recognition in this framework perform better than analogous PLP and MFCC classifiers 
below $18$dB SNR. A combination of the high-dimensional and MFCC features at the likelihood level  performs uniformly better than either of the individual representations across all noise levels.
\end{abstract}


\section{Introduction}
\IEEEPARstart{A}{} major problem faced by state-of-the-art automatic speech recognition (ASR) systems is a lack of robustness, 
manifested as a substantial performance degradation due to common environmental distortions, due to a discrepancy between training and run-time conditions, or due to spontaneous conversational  pronunciation \cite{glass,glass2}.
It was long believed that 
context and language modelling would provide ASR with the level of robustness
inherent to human speech recognition, hence
substantial research efforts have been invested in these higher levels
of speech recognition systems. At the same time, the importance of robust recognition of isolated phonemes, 
syllables and nonsense words has not been fully investigated, while
it is well known that humans attain a major portion of their  robustness in speech recognition 
early on in the process, before and independently of context effects \cite{Miller51,Miller55}. Moreover, for 
language and context  modelling to work optimally, 
elementary speech units 
need to be recognised sufficiently accurately.
In recognising syllables
or isolated words, the human auditory system performs above chance
level already  
at $-18$ dB SNR (signal-to-noise ratio) and significantly above it at $-9$ dB SNR 
\cite{Miller51,Miller55}.  Recent more detailed studies show that human speech recognition remains unaffected by noise 
down to $-2$ dB SNR \cite{jont1,jont2}. 
 No automatic speech classifier is able to achieve
performance close to that of the 
human auditory system in recognising such isolated words or phonemes
under severe noise conditions 
\cite{srokabraida}. 
ASR systems deliver top performance owing to sophisticated language models, combined with hidden Markov model (HMM) advances \cite{gales}. However,
the problem of robustness of ASR, or the lack thereof,  still
persists, and the underlying concern of robustly recognising isolated
phonetic units remains an important unresolved issues. Hence, in this study we explore a novel approach to representing and modelling speech and
investigate its effectiveness in the context of phoneme classification and recognition.

While there are many reasons for the lack of robustness in ASR, one of the major factors could be the
excessive nonlinear compression  that takes place
in the front-end of ASR systems.
As the first step in all speech recognition
algorithms, consecutive speech segments are represented by 
low-dimensional feature vectors, most commonly  {\sl cepstral} features such as Mel-Frequency Cepstral Coefficients (MFCC) \cite{mfcc}
or Perceptual Linear Prediction (PLP) coefficients \cite{hermansky90}.
 Using low-dimensional features was unavoidable when initially introduced in the 1970s \cite{bishnu}, as it removes non-lexical variability irrelevant to recognition, and enables learning of statistical models from limited data and using very limited computational resources. However, this paradigm, which resulted in a major performance boost a few decades ago, might be a bottleneck towards achieving robustness nowadays, when massive amounts of training data are available and computers are orders of magnitude more powerful. 

In the process
of discarding components of the speech signal
that are considered unnecessary for recognition, 
some information that makes speech such a robust message representation might be lost.
It is commonly believed that speech representations that are used
for compression also provide good feature vectors for speech 
recognition. The rationale is that, since speech can be reconstructed
from its compressed form to sound like natural 
speech that the human auditory system can recognise quite reliably,
then no relevant information is lost due to the compression. 
Speech production/recognition is, however, analogous to a channel coding/decoding
problem, while speech compression is a source coding problem, and these two 
are fundamentally different. In particular,
speech production embeds redundancy in speech waveforms in a highly structured 
manner, and distributions of different phonetic units
can withstand a significant amount of additive noise and  
distortion before they overlap significantly. 
Speech compression and standard ASR front-ends, on the other hand,
remove most of this redundancy in a manner that is optimal from a source coding
perspective, and 
represent speech in a space of a relatively low dimension 
where different speech units, although separated, may not be sufficiently far
apart from each other; they may then overlap considerably already at lower 
noise levels than in the original domain of acoustic waveforms. 
Human speech perception studies have shown explicitly that the information reduction that takes place in the conventional front-ends leads to a severe degradation in human speech recognition performance and, furthermore, that in noisy environments there is a high correlation between human and machine errors in recognition of speech with distortions introduced by typical ASR front-end processing  \cite{peters,meyer2007,PaliwalAlsteris2003,PaliwalAlsteris2006}.
Hence, in this paper we study models of speech in the high-dimensional domain of acoustic waveforms or some
linear transform thereof. An additional benefit from
considering uncompressed waveforms is that modelling of noisy data, given models in quiet, is straightforward.
This is in contrast with cepstral representations where, due to the nonlinearities involved, distributions change considerably both with noise type
and level. This then makes efficient  adaption of speech models to different noise conditions very challenging.

Linear representations have been considered previously by other authors.
Sheikhzadeh and Deng~\cite{deng94} apply hidden 
filter models directly to acoustic waveforms, avoiding artificial frame boundaries and therefore allowing 
better modelling of short duration events. They consider consonant-vowel classification and illustrate the importance
of power normalisation in the waveform domain, although a full
implementation of the method and tests on benchmark tasks 
remain to be explored. Poritz~\cite{poritz}, and later  Ephraim and Roberts~\cite{ephraim}, consider modelling
speech explicitly as a time series using autoregressive hidden Markov models. Their work inspired more recent advances in that direction by
Mesot and Barber~\cite{mesot-barber-07a}  who develop a switching 
linear dynamical systems (SLDS) framework. The SLDS approach exhibited significantly better performance at
 recognising spoken digits in additive Gaussian noise when compared to standard hidden Markov models (HMMs) used in combination with cepstral features; 
 however, it is computationally very expensive even  when approximate inference techniques are used.  
 Turner and Sahani proposed using modulation cascade processes to model natural sounds simultaneously 
on many time-scales~\cite{mcp}, but the application of this approach to ASR remains to be explored. 
In this paper we do not directly use the time series interpretation and impose no temporal constraints 
on the models. Instead, we investigate the effectiveness of the acoustic waveform front-end for robust phoneme 
classification and recognition using Gaussian mixture models (GMMs), as those models are commonly used in 
conjunction with HMMs in practical systems.

To assess the merits of representing speech without information loss and non-linear transformation, without the potentially interfering effects of
segmentation, 
in Section \ref{sec:coreTIMIT} we first develop Gaussian mixture models for fixed-length segments of speech
and use them for phoneme classification in the presence of additive noise.
Next in Section \ref{sec:framework} we investigate the effect of the segment duration on classification error. In the same section we compare the new high-dimensional representation with PLP  features in a phoneme classification task that uses information from entire phonemes. Finally in Section \ref{sec:recognition} we consider phoneme recognition from continuous speech in the standard HMM-GMM framework. 
In all scenarios that we investigate we find that PLP and MFCC features  excel in low-noise conditions while high-dimensional linear representation achieve better results at higher noise levels, starting already at around 18dB SNR.  We then demonstrate that a combination of high-dimensional linear and cepstral features achieves better results than either of the individual representations across all noise levels. Recently Power Normalised Cepstral Coefficients (PNCC) features \cite{pncc} were proposed for robust speech recognition, but we became aware of these features 
at the time of the submission of this paper, hence these features are not considered in this study.

\section{Models of Fixed-Length Segments of Speech}
\label{sec:coreTIMIT}

Generative classifiers use probability density estimates, $p(x)$, learned for each class of the training data.
The predicted  class of a test point, $x$, is determined as the class $k$ with the greatest likelihood 
evaluated at $x$. Typically the log-likelihood, $\mathcal{L}(x) = \mathrm{log}(p(x))$,
 is used for the calculation.  A test point $x$ is thus assigned to one of $K$ classes using the following function:
\begin{equation}
\label{prediction}
  \mathcal{A}^L(x) = \arg\max_{k=1,\ldots,K} \mathcal{L}^{(k)}(x) + \mathrm{log}(\pi_k)~.
\end{equation}
\noindent  The inclusion above of $\pi_k$, 
the prior  probability of class $k$, means that we are effectively maximising the log-posterior probability 
of class $k$ given $x$. In this section we build probabilistic models of fixed-length segments of acoustic waveforms 
of phoneme classes. Each waveform segment $x$ is thus a vector in $\mathbb{R}^d$, where $d$ is the number of time samples in the segment.

\subsection{Exploratory Data Analysis}
\label{sec:pre}
Towards constructing  probabilistic models of high-dimensional speech 
representations in the acoustic waveform domain,  it is of interest to  investigate possible lower dimensional
structures in the phoneme classes. Supposing that such structure exists and can be characterised
then it could be used to find better representations for speech, and to construct more accurate 
probabilistic models. 
We thus initially deployed data-driven methods for learning possible low-dimensional manifolds,  as explored in~\cite{vasiloglou09,zhangzhao08}, including locally linear embedding~\cite{roweis00lle} and Isomap~\cite{spect_methods}. 
Many speech representations, typically some variant of  MFCC or PLP, reduce the dimension of speech signals using non-linear processing. Those methods do not directly incorporate 
information about the structure of the phoneme class distributions, but  model the properties
of speech perception. The involved non-linear processing, however, 
makes exact noise
adaptation very challenging (see Section \ref{sec:noise}). Instead one would aim to
find non-linear low-dimensional structures in the phoneme
distributions, and exploit this information to build better models
that remain defined in the original high-dimensional space. This could
include Gaussian process latent variable models~\cite{gplvm} (GP-LVM), which
require as input an estimate of the dimension of the non-linear
feature space.  However, while we found that intrinsic dimension estimates suggest the existence of low dimensional nonlinear structures
in the phoneme distributions, our investigation also showed that these structures still had sufficiently many dimensions to make it impractical to sample them adequately for ASR purposes \cite{isit11}.
Hence
we  turn to more generic density models.  In particular we will construct generative models in the high-dimensional
 space which do not attempt to exploit any submanifold structure directly. We will see that approximations 
are still required, again due to the sparseness of the data and because of computational constraints.

\subsection{Gaussian Mixture Models}
\label{sec:gen}

Without assuming any additional prior knowledge about the phoneme distributions, we use Gaussian mixture 
models (GMMs) to model phoneme densities. 
For the case of $c$ mixture components this function has the form:
\begin{equation}
p(x) = \sum_{i=1}^c \frac{w_i}{(2\pi)^{\frac{d}{2}} |\bm{\Sigma}_i|^\frac{1}{2}}\mathrm{exp}\Big{[}-\frac{1}{2}(x-\mu_i)^\mathrm{T}\bm{\Sigma}^{-1}_i(x-\mu_i)\Big{]}
\label{gmmpdf}
\end{equation}
\noindent where $w_i$, $\mu_i$ and $\bm{\Sigma}_i$ are the weight, mean vector and covariance matrix of 
the $i^\mathrm{th}$ mixture component respectively. We additionally 
impose a zero mean constraint for models as a waveform $x$ will be perceived the same as $-x$. With this constraint,
the corresponding models represent all information about the phoneme distributions
 via the covariance matrices and component weights.

Reliable estimation of full covariance matrices is challenging even in the case of standard low-dimensional features when the number of components in the mixture becomes large. The problem is even more pronounced in the case of
high-dimensional waveform features, where at 16kHz already  with 70ms segments  the dimension of the feature space becomes $d= 1020$. We therefore considered using density estimates derived from mixtures of probabilistic principal component
 analysis (MPPCA)~\cite{tipping99mixtures,ita08}. This method  produces a Gaussian mixture model where the covariance matrix of each component is regularised
 by replacement with a rank-$q$ approximation, where $q  < d$.
However, even PPCA requires an excessive number of parameters compared to the typical amount of available data \cite{ita08}. Finally GMMs with diagonal covariance matrices were investigated,
a common modelling approximation when training data is sparse,  used also in state-of-the-art ASR systems for modelling distributions of cepstral features. This modelling approach achieved lowest classification error, so in all experiments reported in this paper GMMs with diagonal covariance matrices are used. 

Diagonal covariance matrices will be a good approximation provided the data is presented in a basis where correlations between features are weak.  For the acoustic waveform representation, this is clearly not the 
case on account of the strong temporal correlations in speech waveforms.   The density model used for the phoneme classes  thus becomes:
\begin{equation}
p(x) = \sum_{i=1}^c \frac{w_i}{(2\pi)^{\frac{d}{2}} |\mathbf{D}_i|^\frac{1}{2}}\mathrm{exp}\Big{[}-\frac{1}{2}(x-\mu_i)^\mathrm{T}\mathbf{C}^\mathrm{T}\mathbf{D}^{-1}_i\mathbf{C}(x-\mu_i)\Big{]}
\label{gmmpdf:2}
\end{equation}
\noindent where $\mathbf{C}$ is an orthogonal transformation selected to decorrelate 
the data at least approximately, while $w_i$, $\mu_i$ and $\mathbf{D}_i$ are the weight, mean vector and diagonal covariance matrix of the 
$i^\mathrm{th}$ mixture component, respectively. 
We  systematically investigated 
candidate decorrelating bases derived from PCA, wavelet transforms and DCTs.  Although the optimal basis for decorrelation on the training set is indeed formed by the phoneme-specific principal components, we found  that the lowest test error is  achieved with a DCT basis. 
 Preliminary experiments with acoustic waveforms showed that,
instead of performing a  DCT on an entire phoneme segment, it is
advantageous to separate DCTs into non-overlapping
sub-segments. We systematically investigated different block lengths, and found that best classification results were obtained with
  10 ms blocks, mirroring (except for the lack of overlaps) the frame decomposition of MFCC 
and PLP. For a sampling rate of 16 kHz, the
transformation matrix $\mathbf{C}$ is then block diagonal consisting
of $160\times160$ DCT blocks. 

While the DCT in the density modelling in (\ref{gmmpdf:2}) 
was motivated  by the need to decorrelate the data, and thus make the approximation of covariance matrices by diagonal ones more accurate, one may view  the end result as a representation of speech waveforms via some form of short-time spectra, analogous to what is done towards  extracting cepstral features.
There is however a fundamental difference between the block-DCT transform in (\ref{gmmpdf:2})  and the short-time magnitude spectra used to derive PLP and MFCC features: the former is an orthonormal  transform, that is,  just a rotation of the coordinate system that preserves all the information, whereas the latter is a non-linear transform that incurs a dramatic information loss.  In particular, while the discrete Fourier transform is also an orthogonal transform, retaining only its magnitude -- as PLP and MFCC do -- is equivalent to mapping at each discrete frequency a whole circle in  $\mathbb{R}^2$ to just the value of its radius.

GMMs as given in (\ref{gmmpdf:2}) are also used for the MFCC and PLP features, except that component means are not constrained to be zero, and ${\bf C}$ is chosen to be 
the identity matrix, since these features already involve some form of DCT and are approximately decorrelated.

\subsection{Model Average}
\label{average}
In general, more variability of the training data can be captured with an increased number of 
mixture components; however, if too many components are used over-fitting will occur. The best compromise 
is usually located by cross validation using the classification error on a development set. 
The result is a single value for the number of components required. However, we observed that the optimal number 
of components decreases with SNR, hence
we use an alternative 
approach and take the model average over the number of components, effectively a mixture of 
mixtures~\cite{BQDA}; this gave consistent classification improvements over individual mixtures across all noise levels. Thus, we start from a selection of models parameterised by the number of 
components, $c$, which takes values in  $\mathcal{C}=\{1,2,4,8,16,32,64,128\}$ or subsets 
of it. The entries in this set are uniformly distributed on a log scale to give a good range of model complexity 
without including too many of the complex models. We compute the model average log-likelihood  $\mathcal{M}(x)$ as:
\begin{equation}
\mathcal{M}(x) = \mathrm{log}\big{(}\sum_{c\in\mathcal{C}}u_c\mathrm{exp}(\mathcal{L}_c(x))\big{)}
\label{posterioraverage}
\end{equation} 
\noindent with the model weights $u_c=\frac{1}{|\mathcal{C}|}$ and $\mathcal{L}_c(x)$ being the log-likelihood of $x$ given 
the $c$-component model. Alternatively the mixture weights $u_c$ could be determined from the posterior densities of 
the models on a development set to give a class dependent weighting, i.e.
\begin{equation}
u_c = \frac{\sum_{x\in\mathcal{D}}\mathrm{exp}(\mathcal{L}_c(x))}{ \sum_{d\in\mathcal{C}}\sum_{x\in\mathcal{D}}\mathrm{exp}(\mathcal{L}_d(x))}
\label{posteriorweights}
\end{equation}
\noindent where $\mathcal{D}$ is a development set. 
Preliminary experiments suggested that using those posterior weights only gives a slight improvement over $u_c=\frac{1}{|\mathcal{C}|}$. We therefore adopt those uniform weights 
for all results shown in this paper.

\subsection{Noise adaptation}
\label{sec:noise}
The primary concern of this paper is to investigate the performance of the trained classifiers in 
the presence of additive noise.  
Throughout this study, when noise is present in testing, it is assumed that it can be modelled by a Gaussian distribution and that the covariance matrix is known exactly or as an estimate. This assumption may not be valid for real world scenarios but is a good approximation, providing the noise is stationary at the phoneme level. In practice the noise variance would have to be estimated from the input signal and there are many methods available that provide good estimates of the SNR  
\cite{kim08,tchorz}.  Two noise types are studied: white Gaussian noise, where the variance is known exactly, and pink noise extracted from NOISEX-92 \cite{noisex} data base, which is not Gaussian and therefore tests the robustness of the system in the case where the Gaussian noise assumption does not hold. The distribution of the noise is then estimated via a Gaussian model that is later used for noise adaptation.
Generative classification is particularly suited for achieving robustness
as the estimated density models can capture the distribution of the noise corrupted
phonemes. As the noise is additive in the acoustic waveform domain, signal and noise 
models can be specified separately and then combined exactly by convolution.

We consider the  case where the SNR is  set at sentence-level. All sentences will therefore 
be normalised to have unit energy per sample in quiet and noisy conditions.  Different phonemes within 
these sentences can have higher or lower energies, as reflected in the density models by covariance $\mathbf{D}$ 
with trace above or below $d$, where $d$ is the dimension of the feature vectors. 
The relative energy of each phoneme class
is thus implicitly used during classification.
 
The adaptation to noise of phonetic classes in the acoustic waveform domain is performed by replacing each covariance
matrix $\mathbf{D}$ with $\tilde{\mathbf{D}}(\sigma^2)$:
\begin{equation}
\label{noisespec:2}
\tilde{\mathbf{D}}(\sigma^2)= \frac{\mathbf{D} + \sigma^2\mathbf{N}}{1+\sigma^2}
\end{equation}
where $\mathbf{N}$ is the covariance matrix of the noise transformed by $\mathbf{C}$, normalised to have trace $d$, and
$\sigma^2$ is the noise variance.  
For white noise, $\mathbf{N}$ is the identity matrix, otherwise it is estimated empirically from noise samples.
  In general a full covariance matrix will be required to specify the noise structure. However, with a suitable 
choice of $\mathbf{C}$ the resulting $\mathbf{N}$ will be close to diagonal, and indeed when $\mathbf{C}$ is a segmented DCT
 we find this to be true in our experiments with pink noise. To avoid the significant computational overheads of 
 introducing non-diagonal matrices, we therefore retain only the diagonal elements of $\mathbf{N}$.
 
 The normalisation by $1+\sigma^2$ arises  as follows. 
Considering that sentences are normalised to unit energy per sample, a vector ${x}$ containing $d$ samples from the sentence has squared norm $d$. A vector ${n}$ of  $d$ noise  samples, of variance $\sigma^2$, will  have average squared norm $E(\|{n}\|^2) = d\sigma^2$. Because the noise is assumed Gaussian, it can be shown that the fluctuations of $\|{n}\|^2$ away from its average are small, of relative order $1/\sqrt{d}$, or order $\sqrt{d}$ overall
\cite{matthew_thesis}.
The cross-term in the squared norm of the noisy sentence vector, 
$$\|{y}\|^2 = \|{x} + {n}\|^2 = \|{x}\|^2 + 2{x}^T{n} + \|{n}\|^2$$ 
is easily shown to be also of order $O(\sqrt{d})$ \cite{matthew_thesis}. Thus altogether, 
$\|{y}\|^2 = d + d\sigma^2 + O(\sqrt{d})$, 
so that the energy per sample is
$$\frac{\|{y}\|^2}{d} = 1 + \sigma^2 + O\left(\frac{1}{\sqrt{d}}\right)~.$$ 
For large, $d$ as in our case, normalising ${y}$ to unit energy per sample is therefore equivalent to rescaling by $1/\sqrt{1 + \sigma^2}$. When this rescaling is applied to the noisy phoneme models, it gives precisely the normalisation factor
in (\ref{noisespec:2}).
 
There is no exact method for combining models of the training data 
with noise models in the case of MFCC and PLP features, as these 
representations involve non-linear transforms of the waveforms. 
Cepstral mean and variance normalisation (CMVN)~\cite{CMVN} is an approach
commonly used in practice to compensate noise corrupted features.  This method requires estimates of 
the mean and variance of the features, usually calculated sentence-wise on the test data or with a moving average over a similar time window. We take
this to be a realistic baseline. Alternatively the required statistics can be estimated from a training 
set that has been corrupted by the same type and level of noise as used in testing. 
Clearly both 
approaches have merit. For example, sentence level CMVN requires no direct knowledge of the test 
conditions, and can remove speaker specific variation from the data.  The estimates will be less accurate
and as a consequence it is difficult to standardise all components in long feature vectors obtained by concatenating frames;
instead, we considered standardising frame by frame.
Using a noisy training set for CMVN requires that the test conditions are known so that data can be either collected 
or generated for training under the same conditions. The feature means and variances can be obtained accurately, and
in particular we can standardise longer feature vectors. However, as the same standardisation is used for all sentences, any
variation due to individual speakers will persist.
We found that standardisation on the noisy training set gives lower error rates 
both in quiet conditions and in noise \cite{matthew_thesis}, hence all results for CMVN given below use this method. 

At this exploratory stage, we study
also the matched-condition scenario, where training and testing 
noise conditions are the same and a separate classifier is trained for 
each noise condition. While in practice it would be impossible
to have a distinct classifier for every noise condition, matched conditions are nevertheless useful in our exploratory classification experiments: 
because training data comes directly from the desired noisy speech distribution, then assuming enough data is 
available to estimate class densities accurately this approach provides the optimal baseline for all noise adaptation 
methods~\cite{Gales96robustcontinuous,rose}. 

\subsection{Classification Results}
For all experiments reported in this paper realisations of phonemes were extracted from the SI and SX sentences of the TIMIT database. The training 
set consists of 3,696 sentences sampled at 16 kHz. Noisy data is generated by applying additive  noise at nine SNRs.
Recall that the SNRs were set at the sentence level, therefore the local SNR of the individual
phonemes may differ significantly from the set value, causing mismatch in the classifiers. In total ten testing and
training conditions were run: $-18$dB to $30$dB in $6$dB increments and quiet (Q). Following the extraction of the 
phonemes there are a total of 140,225 phoneme realisations.  The glottal closures are
removed and the remaining classes are then combined into 48 groups in accordance with~\cite{kaifulee,lmgmm}.   
 Even after this combination some of the resulting groups 
have too few realisations. To ensure that the training procedure is stable, the smallest groups with fewer than 1,500 realisations were increased in size by the addition
of temporally shifted versions of the data, {\sl i.e.\ }if $x$ is an example in one of the small training classes then the phoneme
segments extracted from positions shifted by $k=-100$, $-75$, $-50$, \ldots, $75$, $100$ samples were also included for
training.
For the purposes of calculating error rates, some very similar phoneme groups are further regarded as identical, 
resulting in 39 groups of effectively distinguishable phonemes~\cite{kaifulee}. 
PLP and MFCC features are obtained in the standard manner from frames of width 25 ms, with a shift of 10 ms between 
neighbouring frames.  Standard 
implementations~\cite{rastamat} of MFCC and PLP with default parameter values are used 
to produce a 13-dimensional feature vector from each time frame.  The inclusion of dynamic \DD ~features~\cite{furui} increases the 
dimension to 39.  

 For the MFCC and PLP representations,
we  consider in this experiment the five frames closest to the centre of each phoneme, covering 65 ms, and concatenate their feature 
vectors. Results are shown for the representations with and those without \DD, giving feature
vector dimensions of $5\times39=195$ and $5\times13=65$, respectively.
The acoustic waveform representation is obtained by dividing each sentence into a sequence of 10ms non-overlapping frames,
and then taking the seven frames (70ms) closest to the centre of each phoneme, resulting in a 1120-dimensional
feature vector. Each frame is individually processed using the 160-point DCT. 
We present results for white and pink noise
and will see that the approximation using diagonal covariances $\mathbf{D}$ in the DCT basis is sufficient to give 
good performance. 
The impact of the number of frames included in the MFCC, PLP and acoustic waveform representations
is investigated in the next section.

\begin{figure}[t]
\centering
\includegraphics[width=\figwidth]{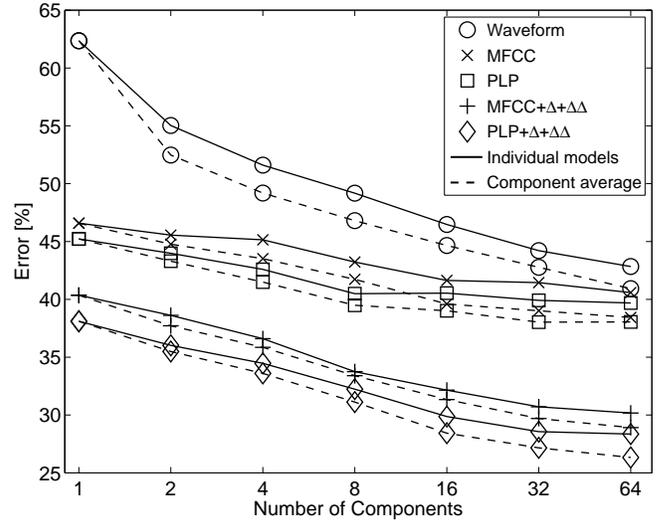}
\caption{\sl Model averaging for acoustic waveforms, MFCC and PLP models, all trained and tested in quiet conditions. 
Solid: GMMs with number of components shown; dashed: average over models up to number of components  shown.  The
 model average reduces the error rate in all cases.}
\label{modelaverage}
\end{figure}

We comment 
briefly on the results for individual mixtures, i.e.\ with a fixed number of components.
Gaussian mixture models were trained with up to 64 components for all representations.  Typically performance on quiet data 
improved with the number of components, although this has significant cost for both training and testing. The optimal
 number of components for MFCC and PLP models in quiet conditions was 64, the maximum considered here. However, 
in the presence of noise the lowest error rates were obtained with few components; typically there was no improvement beyond
 four components.
As explained in Section \ref{average}, rather than working with models with fixed numbers of components, we average over models in all the results reported below. Figure \ref{modelaverage} shows that the improvement
 obtained by this in quiet conditions is approximately 2\% for both acoustic waveforms and PLP with a small improvement
 seen for MFCC also.  The model average similarly improved results in noise.

\begin{figure}[t]
\centering
\includegraphics[width=\figwidth]{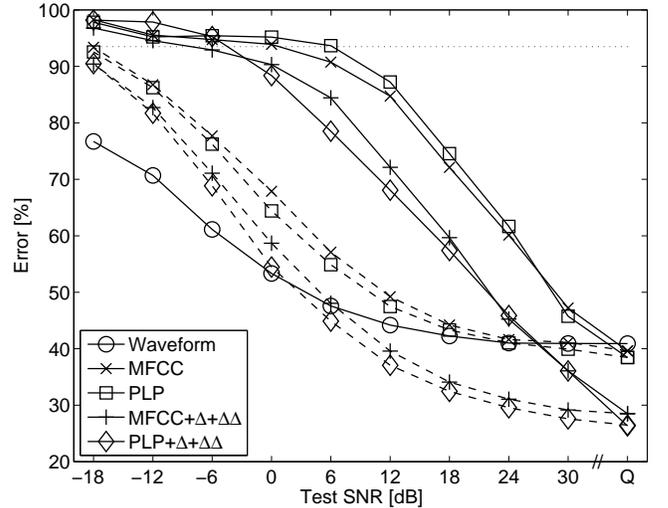}
\caption{\sl Results of phoneme classification using fixed-length segments in white Gaussian noise. All classifiers use the model average of mixtures up to 64 components.
Dotted line indicates chance level at 93.5\%. When the SNR is less that 24dB, acoustic waveforms are the significantly
better representation, with an error rate below chance even at -18dB SNR.  Dashed curves show results of matched 
training for corresponding MFCC and PLP representations. }
\label{adapted}
\end{figure}

One set of key results comparing the error rates in white Gaussian noise for phoneme classification in the three domains is shown in Figure \ref{adapted}.
The MFCC and PLP classifiers are adapted to noise using CMVN. This method is comparable with the 
adapted waveform models as it only relies on the models trained in quiet
conditions.  The curve for acoustic waveforms is for models trained in quiet conditions and then adapted to the appropriate
noise level using (\ref{noisespec:2}). Comparing waveforms first to MFCC and PLP without \DD, we see that in quiet conditions the PLP representation gives the lowest error. 
 The error rates for MFCC and PLP are significantly worse in the presence of noise, however, with acoustic waveforms giving an absolute reduction in error at 0dB SNR of 40.6\% and 41.9\% compared to MFCC and PLP, respectively. 
 Curves are also shown for \MFCCDD\ and \PLPDD,
although these representations include information about underlying speech signals over 145 ms observation windows, {\sl i.e.}
significantly longer than the 70 ms observation windows used to form the acoustic waveform representation.  
Again the same trend holds; performance is good in quiet conditions but quickly deteriorates as
the SNR decreases.  The crossover point is between 24 dB  and 30 dB SNR for both cepstral representations. The chance-level error rate of 93.5\%
can be seen below 0 dB SNR for the MFCC and PLP representations without deltas and below $-6$ dB  SNR when deltas are included, 
whereas the acoustic waveform classifier performs significantly better than chance with an error of 76.7\% even at $-18$ dB SNR.
The dashed curves in Figure \ref{adapted} represent the error rates obtained for classifiers trained in matched conditions. 
The results show that the waveform classifier compares favourably to MFCC and PLP below 24dB SNR when no deltas are appended. This results is in agreement with our working hypothesis that high-dimensional acoustic waveform representations may provide better separation of phoneme classes.
Including \DD\ does reduce the error rates significantly and the
crossover then occurs between 0 dB and 6 dB SNR. It
is these observations that mainly motivate our further model developments
below: clearly we should aim to include information similar to deltas
in the waveform representation.

The same experiment was repeated using pink noise extracted from the NOISEX-92 database~\cite{noisex}. The results for
both noise types were similar for the waveforms classifiers. For \PLPDD, adapted to noise using CMVN, there is a larger difference between the two noise types, with pink noise leading to lower errors.
Nevertheless, the better performance is achieved by acoustic waveforms below a crossover point between 18 dB and 24 dB SNR
\cite{matthew_thesis}. Results for GMM classification on the TIMIT 
benchmark in quiet conditions have previously been reported in~\cite{lmgmm,clarkson} with errors of 25.9\% and 26.3\% respectively. 
To ensure that our baseline is valid we compared our experiment in quiet 
conditions for \PLPDD~and obtained a comparable error rate of 26.3\% as indicated in the bottom right corner of Figure \ref{adapted}.

Following these encouraging results we seek to explore the effect of optimising the number of frames and the inclusion of information 
from the entire phoneme.  The expectation is that including more frames in the concatenation for acoustic waveforms will have a
similar effect to adding \DD\ for MFCC and PLP features. A direct analogue of deltas is unlikely to be useful for
waveforms: MFCC and PLP are based on log magnitude spectra that change
little during stationary phonemes, so that local averaging or differencing
is meaningful. For waveforms, where we effectively retain not just
Fourier component amplitudes but also phases, these phases combine
essentially randomly during averaging or differencing, rendering the
resulting delta-like features useless.

\section{Segment Duration, Variable Duration Phoneme Mapping and Classifier Combination}
\label{sec:framework}

\subsection{Segment Duration}
Ideally all relevant information should be retained by our phoneme representation, but as it is difficult to determine exactly
which information is relevant we initially chose to take $f$ consecutive frames closest to the centre of each phoneme and 
concatenate them. Whilst the precise number of frames required for accurate classification could in principle be inferred 
from the statistics of the phoneme segment durations,
  those durations not only vary 
significantly between classes but also the standard deviation within each class is considerable.  
Since no single value of $f$ will be optimal for all phoneme classes, we  consider the sum of the mixture
log-likelihoods $\mathcal{M}_f$,
as defined in (\ref{posterioraverage}), but now indexed by the number of frames used.  The sum is taken over the set $\mathcal{F}$ 
which contains the values of $f$ with the lowest corresponding error rates, for example $\mathcal{F}=\{7,9,11,13,15\}$ for PLP, giving:
\begin{equation}
\mathcal{R}(\bar{x}) = \sum_{f\in\mathcal{F}}\mathcal{M}_f(x^f)
\label{frameaverage}
\end{equation}  
\noindent where $\bar{x} = \{x^f|f\in\mathcal{F}\}$, with $x^f$ being the vector with $f$ frames. Note that we are adding the
log-likelihoods for different $f$, which amounts to assuming independence between the different $x^f$ in $\bar{x}$. Clearly
this is an imperfect model, as {\sl e.g.\ }all components of $x^7$ are also contained in $x^{11}$ and so are fully correlated, but
our experiments show that it is useful in practice. We also implemented the alternative of concatenating the $x^f$ into
one longer feature vector and then training a joint model on this, but the
potential benefits of accounting for correlations are far outweighed
by the disadvantages of having to fit density models in higher dimensional spaces. Consistent with the independence assumption in (\ref{frameaverage}), in noise we adapt the models
$\mathcal{M}_f$ separately and then combine them as above. The same applies to the further combinations discussed next.

\subsection{Sector sum}

\begin{figure}[t]
\centering
\includegraphics[width=\figwidth]{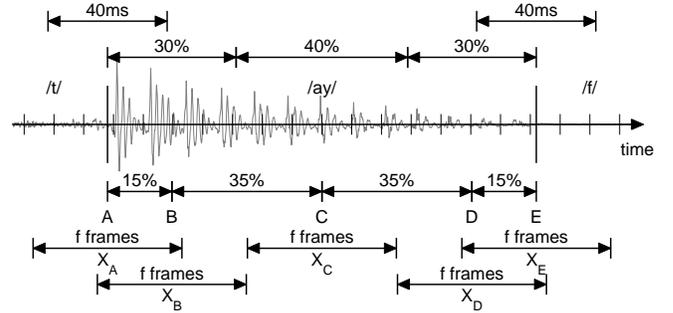}
\caption{\sl Comparison of phoneme representations. Top: Division described in~\cite{clarkson} resulting in five 
sectors, three covering the duration of the phoneme and two of 40ms
duration around the transitions. Bottom: $f$ frames closest to the 
five points A, B, C, D and E (which correspond to the centres of the regions above) are selected to map the phoneme segment to 
five feature vectors 
$x_A$, $x_B$, $x_C$, $x_D$ and $x_E$.}
\label{fig:seg}
\end{figure}

Accounting for the information from the entire phoneme using the standard HMM-GMM framework is considered in the next section in the context of
phoneme recognition. Here we are interested in assessing relative merits of waveform and cepstral representations in robust phoneme classification independently of segmentation errors and interfering effects of HMM assumptions. 
For that purpose, one can consider averaging of frames across the entire phoneme \cite{clarkson} or other methods for mapping variable phoneme duration to fixed-length representation, as  proposed in
\cite{kernel,mari}. Towards GMM models which will be
used in the HMM-GMM framework in the next section, here we
extend the 
centre-only frame concatenation to use information from the entire phoneme by taking $f$ frames with centres closest 
to each of the time instants A, B, C, D and E that are distributed along the duration of the phoneme as shown in Figure \ref{fig:seg}.
In this manner the representation consists of five sequences of $f$ frames per phoneme. Those 
sets of frames are then concatenated to give five vectors $x_A$, $x_B$, $x_C$, $x_D$ and $x_E$. We train five models
 on those sectors and then combine the information they provide by
taking the sum of the corresponding log-likelihoods:

\begin{equation}
\mathcal{S}(\hat{x}) = \sum_{s\in\{A,B,C,D,E\}}\mathcal{M}_s(x_s)
\label{sectorsum}
\end{equation}
\noindent where $\hat{x} = \{x_A,x_B,x_C,x_D,x_E\}$ and $\mathcal{M}_s$ denotes the model for sector $s$, using some fixed number of frames $f$. 
Both improvements can be combined by taking the sum of the $f$-averaged log-likelihoods, $\mathcal{R}_s(\bar{x}_s)$, over the five sectors $s$:
\begin{equation}
\mathcal{T}(\hat{\bar{x}}) = \sum_{s\in\{A,B,C,D,E\}}\mathcal{R}_s(\bar{x}_s)
\label{ssfa}
\end{equation}
\noindent where $\bar{x}_s=\{x_s^f|f\in\mathcal{F}\}$ with $x_s^f$ being the vector with $f$ frames centred on sector $s$, and $\hat{\bar{x}}$ gathers all $\bar{x}_s$. 
Given the functions derived above, the class of a test point can be predicted 
using one of the following:
\begin{equation}
\label{prediction:M}
  \mathcal{A}_f^M(x) = \arg\max_{k=1,\ldots,K} \mathcal{M}_f^{(k)}(x) + \mathrm{log}(\pi_k)
\end{equation}
\begin{equation}
\label{prediction:R}
  \mathcal{A}^R(\bar{x}) = \arg\max_{k=1,\ldots,K} \mathcal{R}^{(k)}(\bar{x}) + \mathrm{log}(\pi_k)
\end{equation}
\begin{equation}
\label{prediction:S}
  \mathcal{A}_f^S(\hat{x}) = \arg\max_{k=1,\ldots,K} \mathcal{S}_f^{(k)}(\hat{x}) + \mathrm{log}(\pi_k)
\end{equation}
\begin{equation}
\label{prediction:T}
  \mathcal{A}^T(\hat{\bar{x}}) = \arg\max_{k=1,\ldots,K} \mathcal{T}^{(k)}(\hat{\bar{x}}) + \mathrm{log}(\pi_k)
\end{equation}
where $\pi_k$ is the prior probability of predicting class $k$.

\subsection{Classification Results}
\label{sec:experiments}
\begin{figure}[t]
\centering
\includegraphics[width=\figwidth]{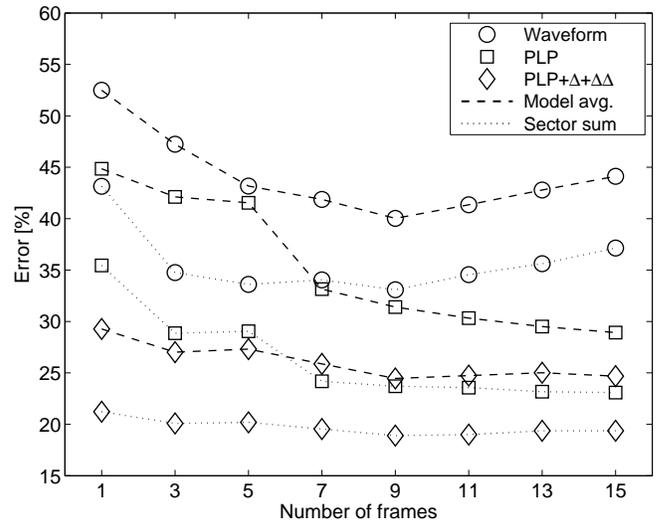}
\caption{\sl Error rates of the different representations in quiet conditions, as a
function of $f$, the number of frames considered. Dashed: prediction
(\ref{prediction:M}) using only the central sector. Dotted: prediction
(\ref{prediction:R}) using the sum over all five sectors, leading to a
clear improvement in all cases.}
\label{fig:frames}
\end{figure}

Figure \ref{fig:frames} shows the impact of the number of frames
concatenated from each sector and the impact of summing log-likelihoods over the five sectors on the classification error, focusing
on quiet conditions. Since PLP features gave slightly lower error than MFCC features in the experiments using centre-only representation, in this section we show results for PLP features only.
We see that the best results for acoustic waveform classifiers are achieved around 9 frames, and 
around 15 frames for PLP without deltas.  The \PLPDD~features are less sensitive to the number of frames with little difference
 in error from 1 to 15 frames. It is worth noting that 
 if we consider the best results obtained for PLP without deltas, 22.4\% using 15 frames, with the best for \PLPDD, 19.6\%
 with 9 frames, then the performance gap of 2.8\% is much smaller than if we were to compare error rates where both classifiers
 used the same number of frames. 
 The error rates 
obtained using the $f$-average over the five best values of $f$ are 32.1\%, 21.4\% and 18.5\% for acoustic waveforms, PLP and \PLPDD,~respectively.
Table \ref{tab:red} shows the absolute percentage error reduction for each of the four
classifiers (\ref{prediction:M})--(\ref{prediction:T}) in quiet
conditions, compared to the GMM with the single best number of mixture components and
number of frames $f$. The relative benefits of the $f$-average and the sector
sum are clear. The sector sum gives the bigger improvements on its own in all
cases compared to only the $f$-average, but the combination of the two
methods is better still throughout.  The same qualitative trend holds true in noise.

Figure \ref{fig:final} compares the performance of the final classifiers, including both the $f$-average and the sector sum, on data corrupted by pink noise. The averaging over the number of frames is done for the five values of $f$ achieving lowest errors, as shown in Figure \ref{fig:frames}.
The solid curves give the results for the acoustic waveform classifier
adapted to noise using (\ref{noisespec:2}), and for the PLP classifier
with and without \DD\ trained in quiet conditions and adapted to noise
by CMVN. The errors are generally significantly lower than in
Figure  \ref{adapted}, showing the benefits of $f$-averaging and the
sector sum. \PLPDD\ remains the best representation for very low noise,
but waveforms give lower errors below crossover points around
 $18$dB SNR. As before, they also perform better than chance down to $-18$dB
SNR.
The dashed lines in Figure \ref{fig:final} show for comparison the
performance of PLP classifiers trained in matched conditions. 
The matched conditions \PLPDD\ classifier 
has the best performance for all SNR. However, in high noise the adapted acoustic waveform
classifier is significantly closer to matched \PLPDD\ than \PLPDD\ with CMVN.
As explained, the CMVN and matched curves for PLP provide the extremes between which we would expect a
PLP classifier to perform if model adaption analogous to that used
with the acoustic waveforms was possible, or some other method to
improve robustness was employed. 
 In the next section, we
explore the performance of MFCC features in combination with   the ETSI advanced front-end
(AFE)~\cite{etsi-afe}
and vector Taylor series~\cite{vts}
noise compensation techniques in the context of phoneme recognition.

\begin{table}[h]
\renewcommand{\arraystretch}{1.3}
\caption{\sl Absolute reduction in percentage error for each of the classifiers (\ref{prediction:M})--(\ref{prediction:T})
in quiet conditions.}
\label{tab:red}
\centering
\begin{tabular}{  l | r | r | r  }
  \hline
  \bf{Model} & Waveform & PLP & \PLPDD \\
  \hline
  Model average ($\mathcal{A}^M$)& 1.6  & 2.8 & 4.4 \\
  \hline
  $f$-average ($\mathcal{A}^R$)& 5.6  &  6.0 & 6.3\\
  \hline
  Sector sum ($\mathcal{A}^S$)& 6.7  & 8.4 & 8.7\\
  \hline
  $f$-average + sector sum ($\mathcal{A}^T$)& 9.9  & 10.0 & 10.4\\
  \hline
\end{tabular}
\end{table}

\subsection{Combination of PLP and Acoustic Waveform Classifiers}
\label{sec:rep}
\begin{figure}[t]
\centering
\includegraphics[width=\figwidth]{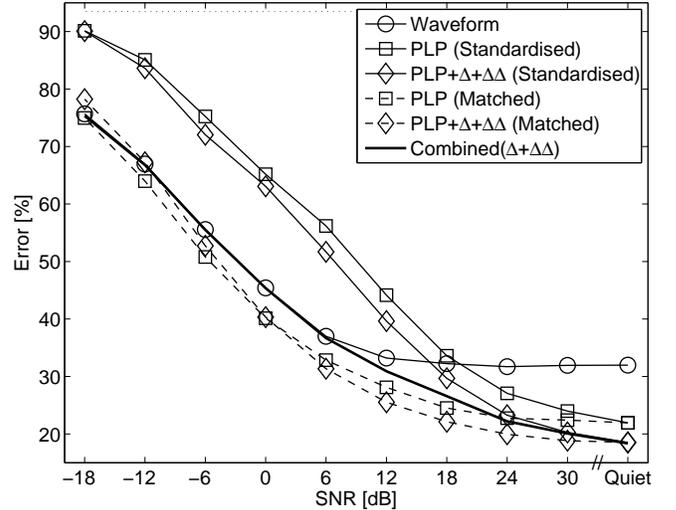}
\caption{\sl Performance of the classifiers in pink noise. Error curves  shown are obtained using the sector sum and averaging over the number of frames for five values of $f$ achieving lowest classification errors according to the results shown in 
 Figure \ref{fig:frames}. Bold line: combined waveform and \PLPDD\ classifier, with the latter adapted to noise by feature standarisation using CMVN.}
\label{fig:final}
\end{figure}

We see from the results shown so far that, as in the preliminary experiments, PLP performs best in quiet conditions with
acoustic waveforms being more robust to additive noise.  To gain the benefits of both representations, we explore 
merging them  via a convex combination of the corresponding log-likelihoods, parameterised 
by a coefficient $\alpha$:
\begin{equation}
\mathcal{T}_\alpha(x) = (1-\alpha)\mathcal{T}_\mathrm{plp}(x) + \alpha\mathcal{T}_\mathrm{wave}(x)
\label{lcomb:2}
\end{equation}
\noindent where $\mathcal{T}_\mathrm{plp}(x)$ and $\mathcal{T}_\mathrm{wave}(x)$ are 
the log-likelihoods of a point $x$ corresponding to PLP and waveform features, respectively. $\mathcal{T}_\alpha(x)$ is then used in place of $\mathcal{T}(x)$ in (\ref{prediction:T}) to
predict the class.  
We would expect optimal $\alpha$ to be almost zero for high SNRs and close to one for low SNRs. In order to achieve the desired improvement in accuracy, we  fit  an appropriate combination function $\alpha(\sigma^2)$. To that end, a suitable range of possible values of $\alpha$ was identified at each noise level from the condition that the error rate is no more than 2\% above the error for the best $\alpha$. This range is broad, so the particular form of the fitted combination function is not critical \cite{ager08}. We choose the following sigmoid function with two parameters $\sigma_0^2$ and $\beta$:
\begin{equation}
\alpha(\sigma^2)=\frac{1}{1+e^{\beta(\sigma_0^2-\sigma^2)}}~.
\end{equation}
A fit through the numerically determined suitable ranges of $\alpha$ then gives  $\sigma_0^2=17$ dB and $\beta=0.3$.
Note that this approach  is  equivalent to a multistream model, where each 
sector and value of $f$ is an independent stream. 
The error of the combined classifier using models trained in quiet conditions is shown as the bold  curve 
in Figure \ref{fig:final}. In quiet conditions the combined classifier is slightly more accurate (18.4\%) than \PLPDD\ alone, corresponding to a small value of $\alpha=0.003$. 
When noise is present the combined classifier is at least as accurate as the acoustic waveform classifier, and significantly better around 18 dB SNR. The combined classifier does improve upon
 \PLPDD\ classifiers trained in matched conditions at very low SNR and narrows the performance gap to the order of 
no more than 9\% throughout, rather than 22\% when comparing to
\PLPDD\ adapted by CMVN.

\section{Phoneme Recognition from Continuous Speech}
\label{sec:recognition}

We now consider extending the classification results from the previous sections to the  task of phoneme recognition from continuous speech using hidden Markov models. 
As GMMs are used for classification as well as for the emission density models of the HMMs, our developments so far can be transferred to recognition. The only exception is the sector sum, which was only intended to mimic the states of HMMs. The model average and frame average, on the other hand, remain suitable for HMMs. Most importantly, the noise adaptation for classifiers in the acoustic waveform domain given by equation (\ref{noisespec:2}) can be directly applied to provide a good model of continuous noisy speech, while the transition matrices of the HMM will remain unaltered in noise based on the assumption that the noise and speech are independent. 

\subsection{HMM Training}
\label{hmmtraining}

In this study we used the Hidden Markov Model Toolkit (HTK)~\cite{htkbook} to model and recognise the phonemes. 
This is a flexible implementation that allows comparing results obtained using the standard representations to   those obtained with  the acoustic waveforms. However, a number of modifications to the standard training procedure are required 
to make the system work properly with acoustic waveform models.

The standard training process implements a flat-start procedure, which begins with  single component GMMs as emission models of HMMs. The first stage initialises all the prototype emission models to the global statistics of the input frames from all sentences and all phonemes. Next the process updates the emission models for each state on a class-wise basis, where the frames from the training set belonging to each class are used to estimate the density of all states for that class. The final stage aims to maximise the training objective function by re-estimating all parameters of the HMM including the emission models and state transition matrices. These models are then further improved by splitting each emission model component to double the number of components. This process replaces each component by two new components, where the means of each component have been moved apart along the axis of maximum variance in the parent component. Following a split, a number of 
re-estimation iterations are performed. The splitting and re-estimation stages are repeated until each emission model consists of 128 components. Beyond this number of components, the error rates increase due to over-fitting.

When using HTK with acoustic waveforms, the training objective needs to be selected first. There is a choice of maximum likelihood estimation, maximum mutual information, minimum phone error or large margin methods  \cite{lmgmm}. In order to provide the best adaptation to noise, maximum likelihood was used as this gives models that should best reflect the true distribution of the phonemes.
 The other methods may provide better performance in quiet or matched conditions, but it is not clear how optimising the phoneme class models in such manner will affect the noise adaptation, since models trained in this way will not be true generative models of the observed speech and so the noise adaptation of (\ref{noisespec:2}) may fail to produce an accurate model of the  noisy speech.
 
Additionally, HTK prunes potential alignments with likelihoods that are too low relative to the best sequence. If no alignment for a particular sentence can be found with sufficiently high 
likelihood, then the sentence is excluded from the training dataset as this exclusion will reduce the training time by ignoring training data and alignments that have very low likelihood given the current model parameters. Due to the higher dimensionality of the feature vectors for the acoustic waveform representation, the likelihood of potential alignments can cover a much wider range than it does in the case of standard cepstral representations. When the default pruning parameter value -- indicating the allowable tolerance on likelihood values -- is used to train acoustic waveform models, the majority of the data is rejected. Consequently, the pruning  parameter was increased significantly. A small fraction of data was still rejected, but the training process was faster than if the pruning option was disabled.

The training initialisation also needs to be adjusted to ensure convergence towards adequate models for acoustic waveforms. Initially the standard training method with the flat start was used, but 
the results obtained using this  training procedure were poor for acoustic waveforms. We traced the problem to the component splitting stage, where the means of the split Gaussian components are moved apart. For acoustic waveform distributions, however, zero mean GMMs provide better models, since a waveform $x$ and $-x$ are identically perceived.
 Following this, a passive method was considered to enforce the zero mean constraint, by including negative instances of the training data. However, the model components still had means that differed significantly from zero. To overcome this problem, the means were instead constrained to zero after splitting and then an update was applied to all parameters except for the mean vectors. This provided a modest improvement, but it appears that the splitting process is the major issue in training acoustic waveform models as the mean vectors will be displaced.
Ultimately the best results are obtained when the GMMs trained on regions $x_B$, $x_C$ and $x_D$ shown in 
 Figure \ref{fig:seg} are used to initialise the emission models for the three states of each phoneme. This initialisation is repeated for each number of components and therefore avoids any splitting of components as the required number of components is specified during the training of the GMMs. Following the HMM initialisation, the variances, transition matrices and component weights are re-estimated using the standard Baum-Welch method.

While model averaging   according to (\ref{posterioraverage}) is still a valid  concept,  it was not used in the phoneme recognition experiments reported in the next subsection, 
 because in the HMM framework the likelihood of the data must be evaluated on all models simultaneously, and this
would increase the computational load significantly. 
We would expect reductions in error rates if we did implement model averaging for HMMs, but the improvement would be similar for all representations 
as observed for the GMM results and would not lead to different conclusions about comparisons of the representations.

As we saw in the previous section, the number of frames, $f$, used for the acoustic waveform representation was also critical for classification. Again a range of values for $f$ were investigated to find those that are most suitable for recognition. This was necessary as the optimal feature duration of acoustic waveform representations is not known, however the parameters used for the standard representations and the results of the classification experiments of the previous sections provided an initial range for investigation.

\subsection{Recognition results}

\begin{figure}[h]
 \centering
\includegraphics[width=\figwidth]{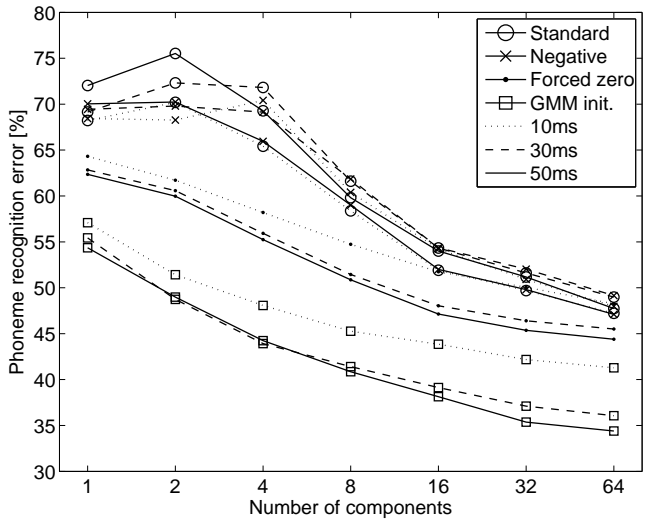}
\caption{\sl  
Comparison of the four training approaches for the acoustic waveform representation in quiet conditions: default training procedure with component splitting (Standard), default with negative instances added to the training data (Negative), default with emission model mean vectors forced to zero after splitting (Forced zero), and emission models initialised using the zero-mean sector models from Section \ref{sec:framework} (GMM init.). Three curves are shown for each of the methods, corresponding to different number of frames in the concatenation.
}
\label{fig53}
\end{figure}

Figure \ref{fig53} shows the phoneme recognition error rates for HMMs using acoustic waveforms, for the four 
training-initialisation approaches detailed in Section \ref{hmmtraining}. This investigation considers the performance of HMMs, when 1, 2, 4, 8, 16, 32 and 64 component GMMs are used as emission density models. Evidently, the standard training procedure is ineffective for acoustic waveforms.  Notice that the inclusion of negative instances of the training data fails to improve the outcome. In both cases,  direct inspection of the model parameters revealed that the models contained components with mean vectors that are significantly different from zero. To overcome this issue, the next method forces the means to zero after each stage of component splitting. This gives much better performance when few components are used. However, the improvement is minor when the models consist of 64 components.
The final training method, denoted as {\sl GMM init.\ }in the legend of Figure \ref{fig53}, initialises the emission models using the GMMs trained in Section \ref{sec:framework}. In this case the error rates are significantly lower than all of the other methods for every number of components. This initialisation method reduces the error rate, ranging from 54.3\% with a single component to 34.4\% with 64 components.

\begin{figure}[h]
 \centering
\includegraphics[width=\figwidth]{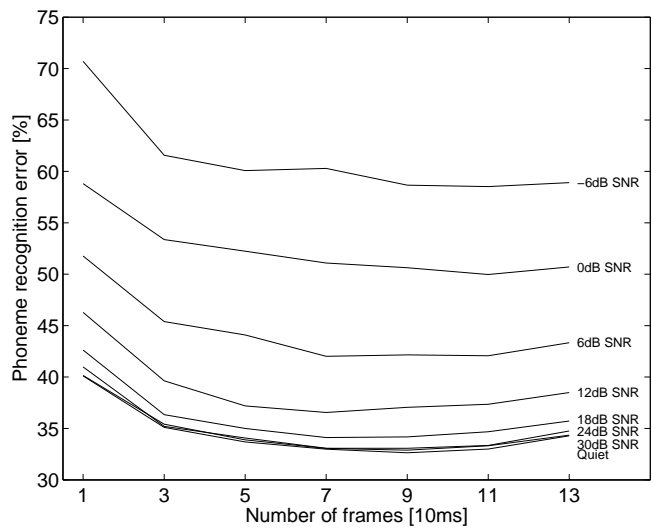}
\caption{\sl  The effect of representation duration, denoted by number of 10 ms frames concatenated. Curves are shown for quiet conditions, and in additive pink noise  for seven values of SNR. Overall the best performance is obtained using 9 frames, a 90ms duration representation. }
\label{fig54}
\end{figure}

The model training process for acoustic waveforms was initially considered for $f = 1$, $f = 3$ and $f = 5$ frames, corresponding to feature vectors covering $10$ ms, $30$ ms and $50$ ms in duration, respectively, extracted every 10 ms. The duration of $10$ ms corresponds to no overlap of adjacent feature vectors, while $30$ ms is closest to the standard cepstral representation. The
effect of the number of frames  is also illustrated in Figure \ref{fig53}. It can be observed that the results improve when going from $f = 1$ to $f =3$ and $f =5$, with no indication that $f =5$ is the optimal value. This suggests that it is necessary to investigate the acoustic waveform representation using more than five frames to find the optimum. In the previous section  representations up to $f = 15$ were considered for classification, and this range was sufficient to find the optimal value of $f$. Figure \ref{fig54} shows the results for the phoneme recognition task over the range from $f=1$ to $f=13$. Each curve represents one fixed global SNR of additive pink noise. There is a slight improvement of around 1.1\% that results from increasing the acoustic waveform duration from 50 ms to 90 ms. Beyond this duration, the error rates increase, which could be due to the representation becoming less localised. 

\begin{figure}
 \centering
\includegraphics[width=\figwidth]{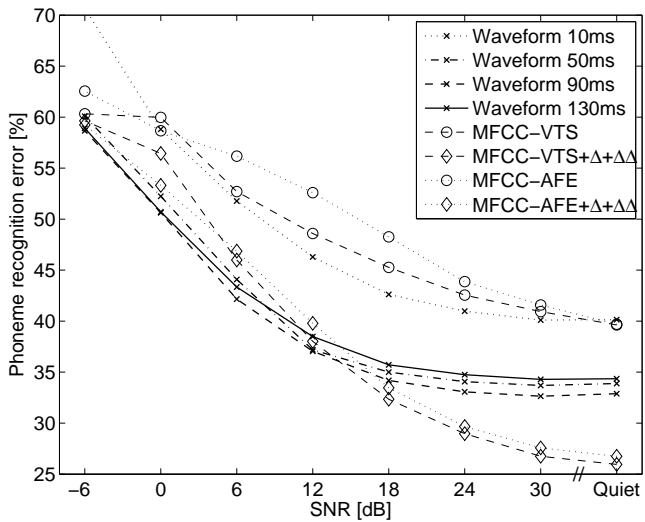}
\caption{\sl  Phoneme recognition error rates for the acoustic waveform representations of different durations.
Tests are carried out in additive pink noise. Noise compensated MFCC features using vector Taylor series (MFCC-VTS) and ETSI advanced front-end (MFCC-AFE) methods are used as baselines.}
\label{fig55}
\end{figure}

A comparison between acoustic waveforms and cepstral baselines is shown in Figure \ref{fig55}.
The models for acoustic waveforms are adapted to additive noise according to (\ref{noisespec:2}), while
in the case of cepstral features, we now consider standard noise compensation techniques: the ETSI 
advanced front-end~\cite{etsi-afe} and vector Taylor series~\cite{vts}, in conjunction with MFCC features to obtain noise-compensated cepstral representations
MFCC-AFE and MFCC-VTS, respectively. 
The compensated features are then standardised using sentence-wise cepstral mean and variance normalisation (CMVN) as that significantly improved performance \cite{matthew_thesis}.  
A single frame is used for MFCC and its variants, with \DD~ computed in the usual manner. This corresponds to a time interval of 25 ms for the frame, 105 ms for $\Delta$  and 145 ms for \DD. It was assumed that these methods provide optimal baselines for comparison as they are commonly used and have been tuned for the task.
Figure \ref{fig55} shows recognition results obtained with the acoustic waveform representation for 10 ms, 50 ms, 90 ms and 130 ms. Considering the comparison to the cepstral baselines without \DD, the acoustic waveform representation with 10 ms duration gives similar error rates to both cepstral baselines in quiet conditions and better performance down to 0 dB SNR. Note that a similar observation can be made about classification results obtained with fixed-length segments of acoustic waveforms and cepstral features without \DD~ information, as shown in Figure \ref{adapted}.
Results shown in Figure \ref{fig55} demonstrate  that in phoneme recognition too the addition of \DD~ features provides  a significant  performance gain for cepstral representations. We can observe again that 
the performance of cepstral representations with \DD~ information is significantly better than that achieved with acoustic waveforms at low SNRs, but that the 90 ms acoustic waveform representation improves further in noisy conditions, giving lower error rates than MFCC-AFE+\DD~ and MFCC-VTS+\DD~ between $-6$ dB and 12 dB SNR.

\begin{figure}[t]
\centering
\includegraphics[width=\figwidth]{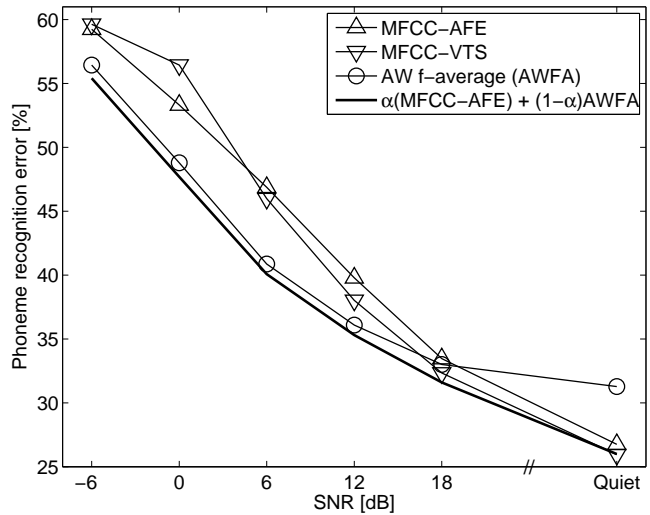}
\caption{\sl Comparison of phoneme recognition error rates in additive pink noise.  Baseline comparisons are provided by MFCC-AFE and MFCC-VTS.   The $f$-average (AWFA) is taken for the range 50ms - 130ms. Uniformly lower error rates are achieved by the convex combination  of MFCC-AFE and AWFA.}
\label{fig:recognition}
\end{figure}

We also re-investigate in the HMM context the effect of 
model averaging by combining models corresponding to different number of  frames used to represent acoustic waveforms.
Having established that the lowest  error rates are achieved using $f=9$ frames, the models corresponding to $f \in \mathcal{F}=\{5,7,9,11,13\}$ are averaged by a multi-stream process~\cite{htkbook} and used for recognition. The resulting acoustic waveform $f$-average (AWFA) model achieved better performance than the individual models with a fixed value of $f$. Finally, we re-investigate in the recognition context combining acoustic waveform and MFCC-AFE HMMs  according to (\ref{lcomb:2}).
Note again that the larger $f$-values for waveforms effectively incorporate information that is provided for MFCCs via $\Delta+\Delta\Delta$, which is why we do not consider similar $f$-frame concatenations for MFCCs. 
Phoneme recognition results obtained with the frame averaging and representation combination in additive pink noise are shown in Figure \ref{fig:recognition}.  
The acoustic waveform representation with averaging over the number of frames (AWFA) 
gives lower error rates than the cepstral representations for SNR below a cross-over point around 18 dB.  
Error rates achieved by the combined representation  are uniformly lower than any of the individual representations considered. As a consequence, the combined HMM model achieves significant reductions in error rates compared to cepstral-only models with standard noise compensation techniques.

\section{Conclusion  and Discussion}

In this paper we have explored the potential of improving the robustness of ASR to additive noise by posing the problem directly in the space of acoustic waveforms  of speech, or some linear transform thereof which does not incur any information loss.
In order to assess the separation of phonetic classes in the acoustic waveform domain and the domains of standard cepstral features, we first considered  classification of phonemes using observations of a given fixed duration. A remarkable result of that experiment is
that phoneme classification in the acoustic waveform domain significantly outperforms classification in the cepstral domains already at very low noise levels, and that even at an SNR as low as $-18$ dB it achieves accuracy distinctly above that of random guessing (see Figure \ref{adapted}). A significant performance gain in classification using cepstral features is achieved via \DD~features, which reflect temporal dynamics of speech. While finding analogous features in the acoustic waveform domain is a nontrivial open problem, considerations in Section \ref{sec:framework} demonstrated that a performance gain in the acoustic waveform domain can be achieved by extending the length of the basic observation unit and thus  including the information on the temporal dynamics implicitly.
Next, acoustic waveforms were considered in the context of phoneme recognition from continuous speech. To allow a direct comparison with the cepstral features, and compatibility with existing technologies, the recognition was implemented in  the standard HMM-GMM framework. Note however that \cite{morgan}: {\sl
"The HMM and frame-based cepstra have co-evolved as ASR system components and hence are very much tuned to each other. The use of HMMs as the core acoustic modeling technology might obscure the gains from new features, especially those from long time scales; this may be one reason why progress with novel techniques has been so difficult. In particular, systems incorporating new signal processing methods in the front end are at a disadvantage when tested using standard HMMs."} Still
phoneme recognition using acoustic waveforms achieved lower error rates than cepstral features for noise levels below 18 dB SNR.  

Recognition and classification results presented in this proof-of-concept study demonstrate that representing speech using high dimensional linear features, which incur no loss of information, is a promising direction towards achieving the long sought-after robustness of ASR systems. Significant improvements can potentially be made by more sophisticated modelling of speech in the acoustic waveform domain. The choice of decorrelating basis is another area that opens up scope for further work. 
In these directions  methods such as MLLT \cite{mllt} and semi-tied covariance matrices \cite{semitied} may lead to improved results. We conducted some preliminary experiments with semi-tied covariance matrices 
 in the context of phoneme recognition. This improved recognition performance   in quiet,  but suffered from increased error levels in noise. However, more extensive investigations may prove fruitful. 
Even within the models introduced in this study, given the wide range of possibilities for choice of  parameters, many of these have not been extensively tuned. This includes the weighting of the model averaging over the number of components in GMMs,  and  the averaging over the number of frames  $f$,  either in a class-independent manner, or tuned even further to vary with the phoneme or phonetic group being evaluated;  a uniform weight was assumed throughout (averaging over the number of components was not explored at all in the recognition experiment), but it is likely that the information content used for discrimination is not distributed in exactly this way. Then models could potentially also be developed 
to explicitly model correlations between feature vectors obtained for
different number of frames $f$.

Towards other classes of models, Gaussian scale mixture models (GSMMs) are a subset of GMMs that seem particularly suitable for modelling  the distributions of the speech signal. This is motivated by the observation in our experiments that the covariance matrices of some of the components were approximately scaled copies of each other. 
Densities of the acoustic waveform representation can thus be modelled using mixtures of scaled components where again the means are constrained to zero and the covariance matrices are constrained to be diagonal. The probability density function of the GSMM is thus given by:
$$p(x)= \sum_{i=1}^c \sum_{j=1}^S w_i v_j {\cal N}(x; 0, s_j^2\mathbf{C}^{-1}\mathbf{D}_i (\mathbf{C}^\mathrm{T})^{-1})$$
where the symbols are as in (\ref{gmmpdf:2}), with the addition of $S$ scales and scale weights,
$s_j$ and $v_j$, respectively.  Here, the range of scales corresponds to a model of the amplitude distribution for a given phoneme.  The eight scale model with 128 components provided an error rate that improved by 4.2\% in quiet, compared to the standard 128-component GMM when used for phoneme classification as considered in Section \ref{sec:coreTIMIT}
\cite{matthew_thesis}. 
Further reductions in the error rates may be expected if more scales are used, given that some of the scale distributions are not adequately modelled with only eight scales.
Although these GSMM models achieved lower error rates  than the corresponding GMMs with the same number of components, computational constraints prevented us from exploring this direction further.
Note that GSMM models are an example of a Richter distribution, which have been previously used
to model heavy tailed distributions \cite{richter}.

\begin{figure}[h]
 \centering
\includegraphics[width=\figwidth]{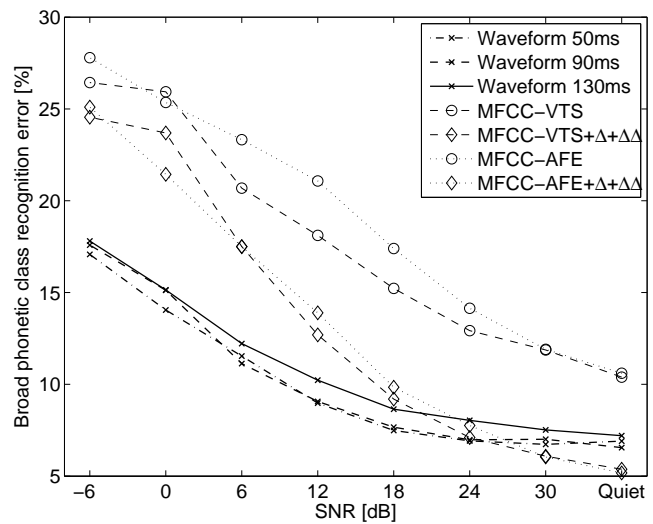}
\caption{\sl Broad phoneme class (vowels, nasals, strong fricatives, weak fricatives, stops and silence) recognition error rates.}
\label{classconfusions}
\end{figure}

We expect that the results can be further improved, and that acoustic waveform features may  by 
effective if incorporated into techniques considered  by
other authors, in particular, committee classifiers and the use of a hierarchy 
to reduce broad phoneme class confusions~\cite{chang}, \cite{broad}. Indeed, in the context of phoneme recognition,
much lower confusion rates between broad phoneme classes are achieved using acoustic waveforms than cepstral features,
as illustrated by results shown in Figure \ref{classconfusions}. 
Finally considering the success of deep neural networks (DNN) in ASR tasks \cite{hinton}, it is imperative to explore DNN modelling of acoustic waveforms of speech. While DNNs  
have exhibited robustness to modest differences between training and test settings, their performance is poor  if the mismatch is more severe \cite{seltzer}. A recent study by Mitra {\sl et al.} \cite{vikram} shows, however, that the choice of features provides scope for improving the  robustness of DNNs. Hence it is of interest to explore possible gains achievable by using DNNs in conjunction with high dimensional linear representations as considered in this study.

\label{sec:discussion}

\section*{Acknowledgment} 
Zoran Cvetkovic would like to thank Jont Allen, Bishnu Atal and Herve Bourlard for their encouragement, support and inspiration, and Malcolm Slaney for his valuable suggestions about the presentation of this paper.

\bibliographystyle{IEEEtran}
\bibliography{refs}
\end{document}